\documentclass[11pt]{article}
\usepackage{booktabs}
\usepackage{amsmath}

\usepackage{xurl}

\usepackage[final]{acl}
\usepackage{tabularx}
\usepackage{makecell}
\usepackage{multirow}
\usepackage{times}
\usepackage{latexsym}
\usepackage[T1]{fontenc}
\usepackage[utf8]{inputenc}
\usepackage{microtype}
\usepackage{inconsolata}
\usepackage{graphicx}
\usepackage{listings}

\usepackage{tikz}
\usetikzlibrary{shapes, positioning, arrows.meta, fit, calc}

\title{Aligning Backchannel and Dialogue Context Representations via Contrastive LLM Fine-Tuning}

\author{
  \textbf{Livia Qian} \and 
  \textbf{Gabriel Skantze}
\\
  Department for Speech, Music and Hearing
\\
  KTH Royal Institute of Technology, Sweden
\\
  \href{liviaq@kth.se}{liviaq@kth.se}, \href{skantze@kth.se}{skantze@kth.se}
}

\begin{document}
\maketitle
\begin{abstract}
Backchannels (e.g., `yeah', `mhm', and `right') are short, non-interruptive feedback signals whose lexical form and prosody jointly convey pragmatic meaning. While prior computational research has largely focused on predicting backchannel timing, the relationship between lexico-prosodic form and meaning remains underexplored. We propose a two-stage framework: first, fine-tuning large language models on dialogue transcripts to derive rich contextual representations; and second, learning a joint embedding space for dialogue contexts and backchannel realizations. We evaluate alignment with human perception via triadic similarity judgments (prosodic and cross-lexical) and a context–backchannel suitability task. Our results demonstrate that the learned projections substantially improve context-backchannel retrieval compared to previous methods. In addition, they reveal that backchannel form is highly sensitive to extended conversational context and that the learned embeddings align more closely with human judgments than raw WavLM features.
\end{abstract}

\section{Introduction}
\label{sec:intro}

Conversational feedback refers to short, non-interrupting responses signaling, e.g., attention, understanding, and surprise~\cite{10.1093/jos/9.1.1}. These responses streamline communication by establishing \textit{common ground}~\cite{CLARK1989259, fusaroli2017measures} and replacing explicit metalinguistic exchanges — for instance, substituting full answers to the question ``Did you understand?'', such as ``Yes, continue'' or ``No, please repeat'', with simple tokens like ``yeah!'' or ``sorry?'', respectively. Feedback is typically multimodal, involving vocalizations, gaze, and gestures~\cite{bertrand2007backchannels, truong2011multimodal, ferre2017unimodal}. Modeling these signals is crucial for building rapport in conversational systems~\cite{axelsson2022modeling}. Vocal instances, such as `yeah', `uh-huh' or `wow!', are commonly termed \textit{backchannels}~\cite{Yngve70}.

\begin{figure}
  \centering
    \includegraphics[width=1.0\columnwidth]{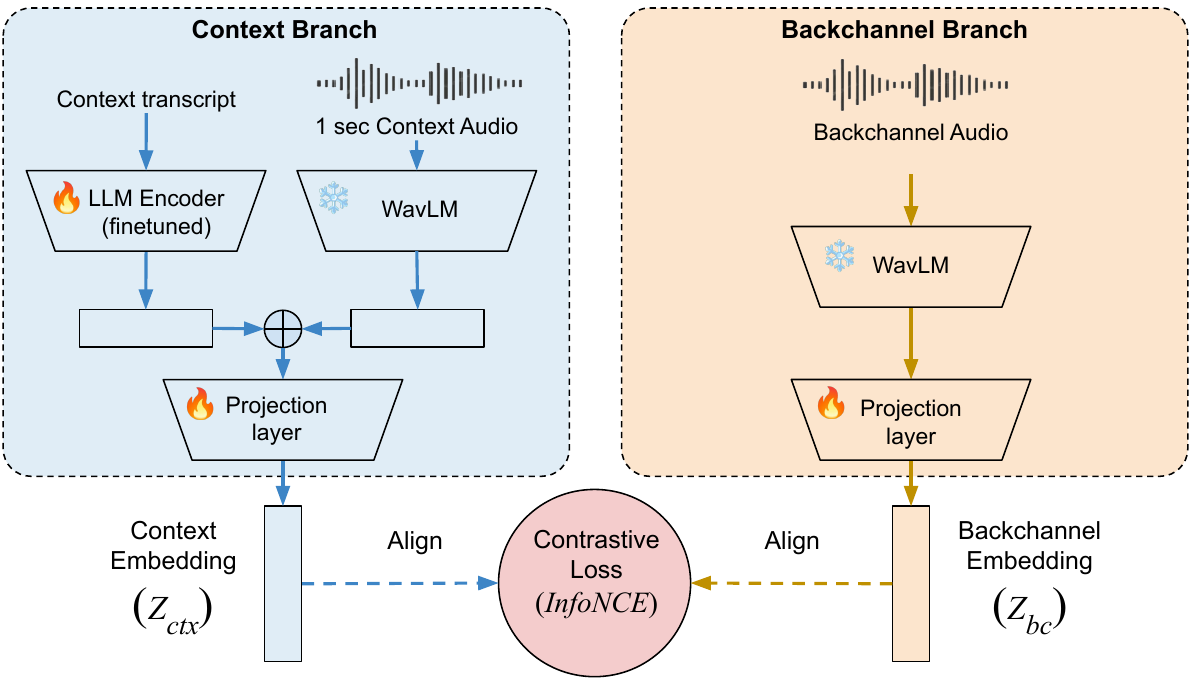}
  \caption{Architecture of the joint context-backchannel model. In this version of the model, both the context transcript and the context audio are passed through their respective encoders (fine-tuned LLM and pre-trained WavLM) before being concatenated and projected to a lower-dimensional space to form a context embedding.
  }
  \label{fig:architecture}
\end{figure}

Computational work has prioritized backchannel placement and timing~\cite{heldner2013backchannel, ruede2019yeah, DBLP:conf/icassp/OrtegaLV20, 10.1145/3472306.3478360} but the link between \textit{form} and \textit{meaning} is often overlooked. Prior work~\cite{benus2007prosody} shows that lexical choice and prosody both shape the pragmatic interpretation of backchannels — e.g., when distinguishing `yeah!' from `yeah?'. Consequently, inadequate use of backchannel forms risks resulting in pragmatically inappropriate responses.

To use and interpret backchannels effectively, robust representations are required. One promising approach embeds backchannel signals in a continuous space where distance reflects similarity. Our earlier work~\cite{liviaqa} showed that this is feasible using a contrastive learning framework that projects past context and feedback into a shared space. However, that work relied on simplistic mean-pooled text embeddings (e.g., BERT by ~\citealp{bert}) to encode dialogue context, limited to the last 4 seconds of the previous turn. 

We extend this method by fine-tuning an autoregressive large language model (LLM) on dialogue transcripts (Section~\ref{sec:fine_tuning_llm}); the hidden state preceding a backchannel serves as a dense semantic representation of the context. This is fused with a WavLM~\cite{9814838} encoding of the last second of the corresponding speech (Section~\ref{sec:joint_learning}). The resulting architecture is shown in Figure~\ref{fig:architecture}.

Our contributions are fourfold. First, we show that fine-tuning an autoregressive LLM on spoken dialogue data substantially improves context encoding, which is needed for effective backchannel representations. Second, we highlight the importance of context length for these representations, indicating that the choice of backchannel form is a pragmatically complex phenomenon. Third, we bridge model and human semantics via downstream evaluation tasks grounded in perception data, showing that the learned representations align with human perception. Finally, we demonstrate that interpretable affective dimensions — Energy, Surprisal, and Polarity — can be recovered from the learned representations through linear projections.

\section{Related work}
\label{sec:related}

Research on conversational feedback and backchannels has historically focused on placement and timing, utilizing concepts such as ``feedback relevance spaces''~\cite{heldner2013backchannel, 10.1007/s10849-020-09328-1} and identifying backchannel-inviting cues~\cite{GRAVANO2011601}. While early approaches relied on rule-based acoustic feature extraction~\cite{koiso1998analysis, bertrand2007backchannels, heldner10_interspeech, poppe2010backchannel}, more recent computational work has shifted toward neural methods, diversifying the potential tasks that can be solved, i.e., those based on backchannel form and meaning. Neural methods include prediction and classification tasks~\cite{ruede17_interspeech, 10447196, park-etal-2024-improving, inoue-etal-2025-yeah, fukunaga2025backchannel}, contrastive learning frameworks~\cite{liviaqa}, textless generation~\cite{mai2025realtimetextlessdialoguegeneration}, and multi-task learning approaches~\cite{jang-etal-2021-bpm, liermann-etal-2023-dialogue}.

Recent efforts have also targeted the collection, annotation, and analysis of backchannel data. \citet{10.1007/978-3-030-83527-9_46} labeled feedback attributes (expectedness, valence, and specificity); \citet{figueroa-etal-2022-annotation} provided annotations of feedback functions for Switchboard~\cite{switchboard}; \citet{muller2022multimediate} focused on backchannels in multimodal groups; and \citet{lin-etal-2025-predicting} created a tri-modal dataset with backchannel and turn shift labels. Studies have also examined relative feedback perception~\cite{qian25b_interspeech} and virtual agent backchanneling behavior~\cite{poppe2013perceptual}.

Transformer-based large language models (LLMs) have been shown to capture long-range context and pragmatic cues essential for turn-taking \cite{ekstedt-skantze-2020-turngpt} and response generation~\cite{zhang-etal-2020-dialogpt}.
While modern LLMs power sophisticated chatbots (OpenAI's  \href{https://chatgpt.com}{ChatGPT}, Google's \href{https://gemini.google.com}{Gemini}, etc.), these systems typically operate sequentially with clear turn demarcations, often lacking the naturalness required for real-time conversational feedback. Furthermore, these models are text-based and therefore cannot capture the prosodic aspects of contextual cues and backchannel forms. 

In parallel, neural speech models have advanced the representation of prosody, which is critical for pragmatics~\cite{nigelward}. Models like wav2vec 2.0~\cite{baevski2020wav2vec}, HuBERT~\cite{hubert}, and WavLM~\cite{9814838} are trained in a self-supervised manner to learn continuous representations of speech. Recently, generative spoken language models (SLMs) have emerged as systems capable of processing and generating audio directly~\cite{nguyen-etal-2023-generative, defossez2024moshi}, sometimes in a ``full-duplex'' fashion. Some fused SLMs combine text and speech embeddings~\cite{arora2025landscape}. While promising, current SLMs often struggle to generate frequent, semantically controlled backchannels, and it is unclear whether they understand the nuances of backchannels coming from the user~\cite{lin2025full}. 

Our work sits at the intersection of these fields. As the potential space of backchannel forms (considering both lexical choice and prosody) is very large, it is hard to discretize them in a meaningful way for pure next-token prediction. While previous research has shown how contrastive learning can be used to project context and backchannels into a shared space, it relied on simplistic context representations~\cite{liviaqa}. 
Our approach leverages the semantic depth of LLMs to capture the long-range pragmatic constraints of the preceding context while using audio embeddings to capture the prosodic nuances of both the end of the context and the backchannel response, directly addressing the gap between form and meaning in backchannels.

Learned context and backchannel embeddings enable several practical applications. First, cosine similarity between contexts and backchannels can be used to rank candidate responses. Second, aligning the embedding space with interpretable dimensions (e.g., Energy, Surprisal, and Polarity) potentially enables semantic control for backchannel synthesis. Finally, projecting user backchannels into the same space allows for inferring their pragmatic meaning in a dialogue system setting.

\section{Datasets}
\label{sec:datasets}

We use Fisher Part 1~\footnote{\url{https://catalog.ldc.upenn.edu/LDC2004T19},
\url{https://catalog.ldc.upenn.edu/LDC2004S13}}~\cite{cieri-etal-2004-fisher} — a prevalent dataset consisting of 5,850, 3-to-10-minute-long telephone calls between native U.S. English speakers — for training, evaluation, and the perception study, utilizing its time-aligned transcripts for backchannel extraction. The other dataset forming the source of backchannels is FiCa~\cite{figueroa-etal-2024-mhm, fica_dataset}, which is a set of reenacted and spontaneously produced backchannels uttered by a single native English speaker, thereby eliminating speaker-dependent effects. FiCa was used for evaluation and comparison, in the form presented in \citet{qian25b_interspeech}.

The set of backchannels included was chosen based on the frequency of their lexical form in the Fisher corpus: `absolutely', `ah', `cool', `definitely', `exactly', `good', `mhm', `mm', `oh', `okay', `really', `right', `sure', `uh-huh', `wow', `yeah', `yep', `yes'. Unlike \citet{qian25b_interspeech}, we do not include responses in the form of `no' and direct displays of non-understanding (`pardon', `sorry', `what') because these, under certain circumstances, are not regarded as backchannels but rather as full turns, answers, and clarification requests.

\section{Models}
\label{sec:models}

Model training consisted of two stages. First, LLMs were fine-tuned on transcripts of Fisher conversations to learn textual context representations (Section~\ref{sec:fine_tuning_llm}). Second, a joint context–backchannel architecture was fine-tuned, leveraging text and speech for \textit{context} modeling and speech alone for \textit{backchannel} generation (Section~\ref{sec:joint_learning}).

The first stage corresponds to fine-tuning the \textbf{LLM encoder}, as illustrated in Figure~\ref{fig:architecture}, while the second stage involves fine-tuning the \textbf{projection layers}. Although the LLM fine-tuning stage does not capture prosodic variations of backchannels, we assume it learns how different conversational contexts shape expectations over the probability distribution of lexical backchannel forms (which carry distinct semantics). This enables the joint training stage to associate these lexical expectations with a richer representation of both prosodic and lexical realizations.

\subsection{Fine-tuning LLMs for contextual semantic features}
\label{sec:fine_tuning_llm}

In this step, we compared the capability of open-weight, state-of-the-art LLMs to model context semantics and to see how the length of prior context (expressed in \textit{number of turns}) affects this. We compared Gemma 3~\cite{gemmateam2025gemma3technicalreport, gemma3_4b}, LLaMA 3.1~\cite{grattafiori2024llama3herdmodels, meta_llama3_2024}, Qwen2.5~\cite{yang2025qwen2, qwen_2_5} and Mistral~\cite{jiang2023mistral7b, mistral}. All models were fine-tuned for causal language modeling with a fixed set of hyperparameters (batch size = 2, max token length = 1024), using QLoRA (attention dimension = 32, alpha = 64, dropout = 0.05).

We split the 83,047 Fisher transcript snippets into training and test sets of equal size. First, we trained the models on full transcripts in the training set, with each transcript consisting of up to 50 turns, including both backchannel and non-backchannel turns. To evaluate the models, we compared their average perplexity over all backchannels found in the test set by feeding in different numbers of preceding turns (1, 3, and 5). For comparison, this was also applied to the corresponding pre-trained models (with 5 preceding turns). The formatting of the transcripts and the calculations are described in Appendix~\ref{sec:appendix_transcripts}. The results of our experiments are shown in Table~\ref{tab:llm1}.

Average perplexity was computed based on the first token of each backchannel; the relative results were comparable when using a weighted average over all tokens, including those in multi-token words. Perplexity consistently decreased as context length increased, highlighting the importance of rich context for determining backchannel form. The three larger fine-tuned models exhibited similar performance, whereas the fine-tuned Gemma 3 4B yielded slightly higher perplexity, likely due to its smaller size. The pre-trained models showed considerably higher perplexity, demonstrating the necessity of fine-tuning; this was anticipated given the specific formatting of the data.

\begin{table}[t]
\centering
\begin{tabular}{lcccc}
\toprule
 Model & 1 & 3 & 5 & 5 (pre) \\
\midrule
Gemma 3 4B & 11.45 & 9.88 & 9.30 & 54.35 \\
LLaMA 3.1 8B & 10.19 & 8.68 & 8.19 & 32.83 \\
Qwen2.5 7B & 10.37 & 8.87 & 8.40 & 55.33 \\
Mistral 7B & 10.87 & 9.30 & 8.76 & 30.91  \\
\bottomrule
\end{tabular}
\caption{Average perplexity on the backchannels in the test set using 1, 3, and 5 preceding turns as context, across fine-tuned LLMs. For comparison, the perplexity scores by the corresponding pre-trained models are also included (for 5 turns). Perplexity is calculated on the first token of each backchannel word. For more details, see Appendix~\ref{sec:appendix_transcripts}.}
  \label{tab:llm1}
\end{table}

\subsection{Joint contrastive learning of context and backchannel embeddings}
\label{sec:joint_learning}

For contrastive learning, we created a joint architecture that projects context and backchannel vector representations into a shared space, forming \textit{joint embeddings}, as illustrated in Figure \ref{fig:architecture}. For \textbf{context}, we take the final layer's hidden representation that is used by the LLM head to predict the next token (the first token of the backchannel). This is concatenated with the last second of audio from the interlocutor's channel (ending at the onset of the backchannel), encoded with WavLM \cite{9814838} and mean-pooled over its final layer. For ablation, we also compare this with using only the LLM or WavLM embeddings. These embeddings are then projected with an MLP to a context embedding ($Z_{ctx}$). For the \textbf{backchannel} embedding ($Z_{bc}$), we simply used the mean-pooled WavLM encoding of the audio with a linear projection. 

\paragraph{Loss} We use a symmetric InfoNCE-style contrastive loss, similar to the objective introduced in \citet{oord2018representation} and later adopted in a symmetric form by \citet{pmlr-v139-radford21a}. Given a set of $N$ pairs (a batch), the loss maximizes the cosine similarity between the representations of the matching (positive) pairs and minimizes the similarity for the non-matching (negative) pairs:

\begin{equation*}
\label{loss}
\begin{aligned}
\mathcal{L}
&= \frac{1}{2} \Big(\mathcal{L}_{context} + \mathcal{L}_{backchannel}\Big) \\
\mathcal{L}_{context}
&= \frac{1}{N} \sum_{i = 1}^{N} CE (S_{i,:}, i) \\
\mathcal{L}_{backchannel}
&= \frac{1}{N} \sum_{j = 1}^{N} CE (S_{:,j}, j)
\end{aligned}
\end{equation*}

$CE$ is the cross-entropy loss, and $S$ is a temperature-scaled $N \times N$ cosine similarity matrix containing similarity scores between the joint embeddings of all contexts (rows) and all backchannels (columns). Considering that the goal is to maximize the diagonal, i.e., the similarity score of the ground truth pairs, this matrix can also be viewed as logits from the joint model; in this case, the task can be framed as a multiclass classification problem where the number of classes is equal to the number of true pairs ($N$). For each context, the model maximizes the score of the true backchannel among $N$ candidates; the same applies to the backchannels trying to optimize the score of their own contexts. As can be seen, this method is largely affected by the batch size $N$.

\paragraph{Dataset} Using the backchannel forms listed in Section~\ref{sec:datasets}, 105,209 samples, i.e., context-backchannel pairs, were found in the Fisher dataset. Ground truth context-backchannel pairs form positive samples, while all other combinations serve as negative samples. 
Batches (sets of pairs) are shuffled and include multiple speakers for robustness and generalizability. We used an 80-10-10\% train-validation-test split with mutually exclusive speakers and dialogues and ensured that the validation and test data were not used for LLM fine-tuning.

\paragraph{Hyperparameters} With a predefined temperature of 0.07 (as used in~\citealp{pmlr-v139-radford21a}), we tuned the following hyperparameters within each category of \textbf{text embedding} (fine-tuned Gemma, LLaMA, Qwen, and Mistral, as well as the baseline GTE~\cite{li2023towards}): \textbf{the number of layers} in the context encoder's MLP (1, 2, 3, 4), \textbf{embedding size} (64, 128, 256), and \textbf{batch size} (1024, 2048, 4096, 8192). GTE (General Text Embedding) operates with mean-pooled deep contextualized embeddings; it was used as a baseline due to its performance in \citet{liviaqa}. We also found that projecting the feedback with a linear layer is sufficient, and larger MLPs decrease performance due to the simplicity of the feedback embeddings; therefore, no hyperparameter search was needed on the feedback side. We checked the best validation results within $20$ epochs, based on \textit{top-10\%} accuracy (see \textit{Metric} below). The best hyperparameter configurations can be found in Appendix~\ref{sec:hyperparam}.

\paragraph{Metric} Optimizing cosine similarity scores can be viewed as a ranking task; for each context, the joint model provides a list of predictions in order of similarity. As mentioned before, this is equivalent to $N$-class classification, where the classes are the backchannels in the given batch, making accuracy a fitting metric. However, since backchannels are often very similar and batches are large, we consider \textit{top-k} accuracy. Since this metric is relative to batch size, we use \textit{top-k\%} accuracy, similarly to \citet{liviaqa}, which yields a random baseline of $k\%$. In this paper, we report $k=10$.

\paragraph{Best hyperparameters} Models favored mid-to-large batch sizes (2048, 4096) and generally smaller embedding sizes (64, 128), with no consistent pattern in number of layers. This suggests that dense, sometimes shallow embeddings suffice for modeling the context-backchannel relationships in our dataset. The preference for larger batch sizes was expected, as they improve discrimination between positive and negative pairs (\citet{pmlr-v139-radford21a}, for example, used a batch size of 32,768), but the question of why the largest chosen batch size (8192) is relatively underrepresented remains.

\paragraph{Results} The top-10\% test accuracy results are shown in Table~\ref{tab:joint}. Autoregressive LLMs clearly outperform the SSL models in \citet{liviaqa}, where HuBERT combined with GTE achieved 36.45\%. Our GTE baseline performs slightly worse, likely due to a somewhat different selection of backchannels. Consistent with prior findings, combined modalities yield the best representations (with some fluctuations). There were no significant differences among the LLMs, suggesting that small LLMs are sufficient for this problem.

\begin{table}[t]
\centering
\begin{tabular}{lccc}
\toprule
& \multicolumn{3}{c}{\textbf{Context modality}} \\
\cmidrule(lr){2-4}
\textbf{\makecell[l]{Text \\ embeddings}} & Audio + text & Text & Audio \\
\midrule
Gemma 3 4B   &          45.6 &          43.3 & \multirow{4}{*}{29.4} \\
LLaMA 3.1 8B & \textbf{49.8} &          44.2 & \\
Qwen2.5 7B   &          46.3 &          39.7 & \\
Mistral 7B   &          44.7 & \textbf{46.4} & \\
\cmidrule{1-4}
GTE baseline & 33.4 & 21.2 & - \\
random       & 10.0 & 10.0 & 10.0 \\
\bottomrule
\end{tabular}
\caption{Top-10\% test accuracy on the test set, in relation to context embedding modality and text embedding type (irrelevant for the last column). Here, the context consists of 5 turns.
}
\label{tab:joint}
\end{table}

For ablation, we also examined the impact of context length (1, 3, and 5 turns) and LLM fine-tuning, as shown in Table \ref{tab:joint-ctxlen}. These results reflect the perplexities of the LLMs reported in Table \ref{tab:llm1}, reinforcing the finding that LLM fine-tuning is an important first step, and that longer contexts are important to consider in order to predict backchannel forms. The best-performing architecture is henceforth referred to as \textit{the joint model}.

\begin{table}[t]
\centering
\begin{tabular}{lcccc}
\toprule
 Model       & 1    & 3    & 5    & 5 (pre)    \\
\midrule
Gemma 3 4B   & \textbf{40.1} &   40.3 & 45.6 & 40.9\\
LLaMA 3.1 8B &          39.9 & \textbf{46.5} & \textbf{49.8} & \textbf{41.4} \\
Qwen2.5 7B   &          39.7 &   42.4 & 46.3 & 40.9 \\
Mistral 7B   &          37.3 &   43.6 & 44.7 & 41.3 \\
\bottomrule
\end{tabular}
\caption{Top-10\% test accuracy on the test set, for the full (audio + text) model, depending on how many preceding turns were given to the LLM. Pre-trained baselines with a context length of 5 turns are also provided.}
  \label{tab:joint-ctxlen}
\end{table}

\section{Downstream tasks}
\label{sec:downstream}

While the results from the contrastive learning demonstrate internal alignment between context and backchannel embeddings, they do not guarantee that the learned embeddings correspond to human perception of semantic similarity, neither within or across the embedding spaces. To assess external validity, we collected human perception data for downstream evaluation.

\subsection{Human perception data}
\label{subsec:human_perception_data}

First, we used the dataset presented by~\citet{qian25b_interspeech}, where participants had compared triplets of feedback responses with identical lexical forms and without context (from the multi-speaker \textit{Fisher} and the single-speaker \textit{FiCa} datasets), selecting the two most similar in terms of prosody and semantics. Using this dataset, we assessed representational quality by expecting the selected pair to be the closest in the embedding space, measured using cosine similarity. We call this the \textbf{prosodic backchannel similarity task}, as the lexical forms are identical.

We collected another perception dataset using \textit{Fisher}
samples that were not included in the training set for the LLM fine-tuning or the joint model. Participants evaluated a continuous, uninterrupted single-turn context (auditory and textual, 4-80 s) against three audio-only backchannels: \textit{one ground truth} and \textit{two distractors} from different speakers in different dialogues. Unlike the previously mentioned experiment, the candidate responses differed in lexical form. The participants were given three different tasks for each combination of context and three responses, henceforth referred to as \textit{stimulus set} (for details, see Appendix~\ref{sec:appendix_questions}, ~\ref{sec:appendix_questions2} and~\ref{sec:informed}):

\paragraph{Cross-lexical backchannel similarity task} This is a triadic comparison task in which participants select the most similar pair of backchannels. It differs from the \textit{prosodic backchannel similarity task} in that different lexical forms are presented. Here, too, perceived similarity is expected to align with proximity in the embedding space. Results of this task and the \textit{prosodic backchannel similarity task} are reported in Section~\ref{sec:similarity_task}.

\paragraph{Context-backchannel matching task} Participants rate each backchannel's suitability and naturalness in the given context (on a scale of 1-5). We leverage the ground truth to compare human accuracy against the model, evaluating the alignment of context and backchannel embeddings. Results are reported in Section~\ref{sec:matching}.

\paragraph{Affective rating task} Participants rate backchannels (1–5) on \textbf{Energy} (how energetic the response sounds), \textbf{Polarity} (how positive the response sounds), and \textbf{Surprisal} (how surprised the backchannel speaker sounds). These names were chosen to be fairly easy for annotators to understand. Polarity and Energy correspond to the notions of \textit{valence} and \textit{arousal}, commonly used in psychology, psycholinguistics, and affective computing ~\cite{kuppens2013relation, warriner2013norms, 8013713}. Surprisal (the level of surprise) was chosen due to its importance as a backchannel function~\cite{figueroa-etal-2022-annotation} and because it is not clearly captured by the other two dimensions — although they are slightly correlated, as we show later. We evaluate how well linear probes on projected backchannel embeddings predict these dimensions. Results are in Section~\ref{sec:affective}.

In total, we collected 2100 data points (consisting of replies to all three tasks), with 6300 individual ratings of 1260 unique backchannels in 420 unique contexts.

\subsection{Backchannel similarity tasks}
\label{sec:similarity_task}

For the similarity tasks, we selected samples (stimulus sets) with $\ge 80\%$ rater agreement on which backchannels are the most similar. We applied the best joint model's backchannel encoder to obtain embeddings ($Z_{bc}$), identified the most similar pair via cosine similarity, and calculated the agreement rate with human consensus. As a baseline, we use the mean-pooled WavLM embeddings fed into the backchannel projection layer, allowing us to isolate and quantify how much the projection improves the representational space.

The results, shown in Table~\ref{tab:triadic_results}, indicate that the projection indeed makes the space more discriminative for backchannel similarity. In particular, embeddings produced by the joint model’s projection layer yield substantially higher agreement with human judgments than the mean-pooled WavLM baseline, demonstrating that the projection reshapes the acoustic representations in a way that better aligns with perceptual similarity. This suggests that the improvements are not merely due to the underlying pre-trained speech features, but rather to the task-specific structures learned during joint training. Overall, these results support the effectiveness of the backchannel projection in capturing fine-grained similarities that are salient to human listeners.

\begin{table}[ht]
  \centering
  \begin{tabular}{lccc}
    \toprule
    & \multicolumn{2}{c}{\textbf{Prosodic}} & \textbf{Cross-lexical} \\
    \cmidrule(lr){2-3} \cmidrule(lr){4-4}
     & \textbf{Fisher} & \textbf{FiCa} & \textbf{Fisher} \\
    \midrule
    Joint model & 69.7 & 75.9 & 66.3 \\
    WavLM  & 61.7 & 70.8 & 56.6 \\
    Random  & 33.3 & 33.3 & 33.3 \\
    \bottomrule
  \end{tabular}
  \caption{Proportions of correct selections in the triadic backchannel similarity tasks (\%).}
  \label{tab:triadic_results}
\end{table}

\subsection{Context-backchannel matching task}
\label{sec:matching}

For the context–backchannel matching task, we computed the cosine similarity between the context and backchannel embeddings produced by our context and backchannel encoders ($Z_{ctx}$ and $Z_{bc}$), selecting the backchannel with the highest similarity score for a given context. We measured the proportion of cases where the model correctly identified the ground truth backchannel, using 1 or 5 context turns. To compare with human performance, we averaged rater scores for each backchannel option, selecting the backchannel with the highest mean assigned score. We then calculated how often this choice matched the ground truth.

Results (Table~\ref{tab:matching_results}) show that the model, surprisingly, substantially outperforms humans, even when it only has access to 1 context turn (just like humans had in the experiment). An alternative interpretation is that the relatively low human ``performance'' indicates that the same conversational context can allow for multiple valid backchannel responses. Thus, treating the actual backchannel as ground truth should be done with caution.

\begin{table}[ht]
  \centering
  \begin{tabular}{lc}
    \toprule
     \textbf{Rater type} & \textbf{Score}\\
    \midrule
    Joint model (5 turns)& 72.3  \\
    Joint model (1 turn)& 62.0  \\
    Human (1 turn)& 47.4  \\
    Random  & 33.3  \\
    \bottomrule
  \end{tabular}
  \caption{Proportions of correct selections in the context-backchannel matching task (\%).}
  \label{tab:matching_results}
\end{table}

\subsection{Affective rating task}
\label{sec:affective}

\label{sec:affective_dims}

\begin{figure}
  \centering
    \begin{tikzpicture}
        \node (main_image) at (0,0) {\includegraphics[width=1.0\columnwidth]{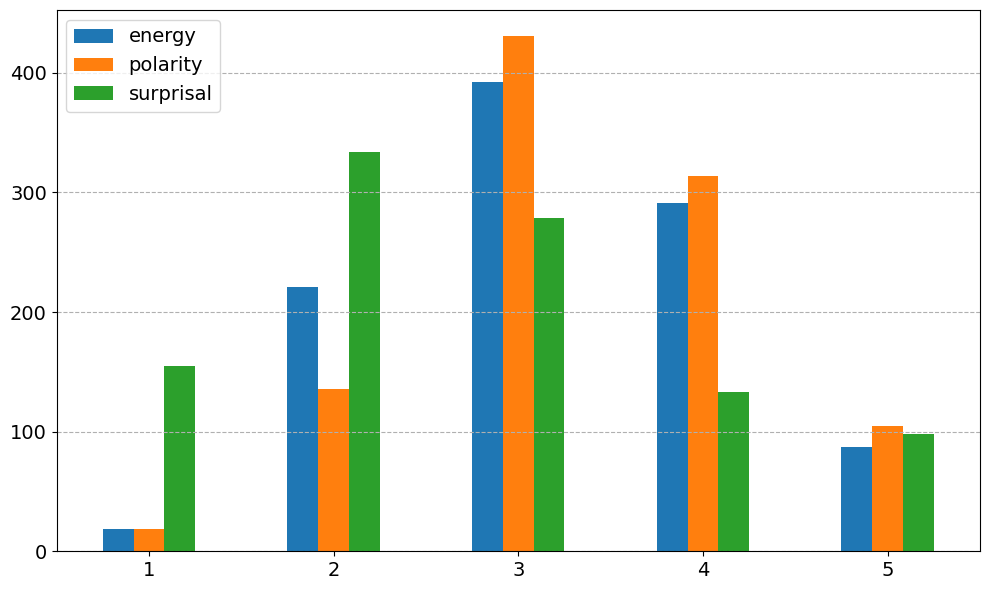}};
        \node at (0.8, 1.0) [
            fill=white, inner sep=1pt, 
            anchor=south west      
        ] {
        \tiny
            \begin{tabular}{|c|c|}
                \hline
                & $R^2$ \\
                \hline
                Energy - Polarity & 0.45 \\
                Energy - Surprisal & 0.39 \\
                Surprisal - Polarity & 0.09 \\
                \hline
            \end{tabular}
        };
    \end{tikzpicture}
  \caption{Distribution of median affective ratings  over backchannels and correlation between mean ratings.}

  \label{fig:affective_scores}
\end{figure}

Figure~\ref{fig:affective_scores} displays the distributions of median ratings for each backchannel: Energy is centered but Surprisal skews lower, which is consistent with how many continuers (e.g., `mhm', `yeah', and `right') do not express surprise. Polarity is predominantly neutral or positive, reflecting the fact that backchannels are less frequently used to express negative sentiments~\cite{jurafsky1998lexical}. The correlation between these dimensions is also shown in Figure~\ref{fig:affective_scores}; this indicates that Surprisal and Polarity correlate the least, while Energy captures some aspects of both. For a more detailed analysis grouped by lexical category, see Appendix~\ref{sec:distribution}.

To measure how well the backchannel embeddings ($Z_{bc}$) capture these dimensions, we split the samples (50/50 train/test) at random and fitted a linear ridge regression probe ($\alpha=1.0$). We compared this against four baselines: raw mean-pooled WavLM embeddings, one-hot encoded lexical tokens, basic prosodic features (\textit{pitch range} in semitones, \textit{duration} in number of voiced frames), and the combination of lexical categories and prosodic features. Pitch was extracted using \href{https://github.com/google/REAPER}{Reaper}.

The results (Table~\ref{tab:affective_results}) show that our learned embeddings capture these perceived dimensions better than WavLM, while simple prosodic features perform the worst. This indicates that the distributed representations integrate lexical and prosodic information more effectively than low-level acoustics or lexical identity alone. Consistent with our earlier findings in this paper, the backchannel projection layer further reshapes the embedding space in a way that improves alignment with human judgments.

\begin{table}[ht]
  \centering
  \begin{tabular}{lccc}
    \toprule
     & \textbf{Energy} & \textbf{Polarity} & \textbf{Surprisal}\\
    \midrule
    Joint model & 0.465 & 0.341 & 0.552 \\
    WavLM  & 0.406 & 0.287 & 0.502  \\
    \midrule
    Lexical  & 0.193 & 0.204 & 0.432  \\
    Prosody  & 0.118 & 0.028 & 0.156  \\
    Lex. + Pros.  & 0.240 & 0.237 & 0.473  \\
    \bottomrule
  \end{tabular}
  \caption{Linear probe fit \big($R^2$\big) on the test set in the affective rating task.}
  \label{tab:affective_results}
\end{table}

\section{Analysis of backchannel embeddings}
\label{sec:analysis}

The downstream tasks demonstrate the utility and validity of the learned context and backchannel embedding spaces. To investigate this further, we developed a tool to visualize backchannel embedding projections in different ways and listen to their prosodic realizations for both the Fisher and FiCa datasets~\footnote{\url{https://qianlivia.github.io/Aligning-Backchannel-and-Dialogue-Context-Representations/}}.

\begin{figure}
  \centering
    \includegraphics[width=1.0\columnwidth]{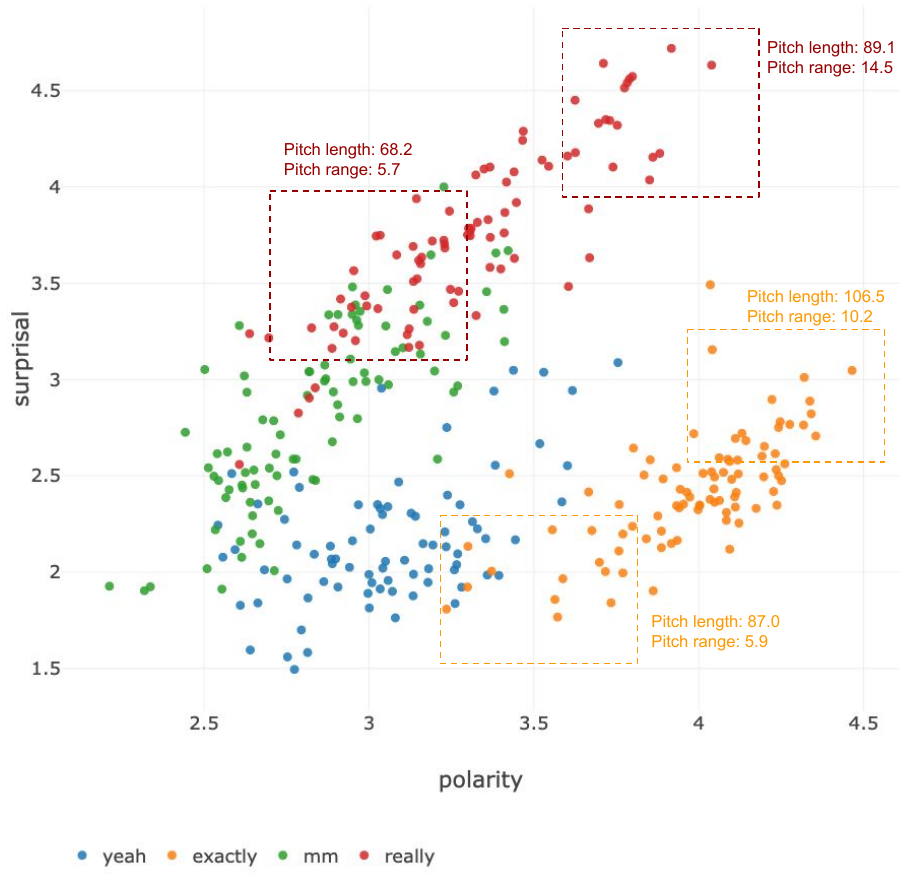}
  \caption{Backchannel samples from the Fisher corpus, embedded through the model, and projected to the Surprisal and Polarity affective dimensions. Rectangles show average pitch length (in voiced frames, 100 Hz) and pitch range (in semitones) for a certain region and for a specific lexical form.}
  \label{fig:pol_sur}
\end{figure}

Figure~\ref{fig:pol_sur} shows a scatterplot of the representation space created by the tool. The backchannel embeddings are projected onto the \textit{Surprisal} and \textit{Polarity} affective dimensions (the least correlated dimensions) using the learned probes from Section~\ref{sec:affective_dims}. For clarity, we only show `yeah', `mm', `exactly', and `really'. The tool can also provide prosodic analysis over a region of backchannels, which is shown as rectangles in the plot. The reported metrics are average duration and average pitch range, as defined in Section~\ref{sec:affective_dims}.

The four lexical tokens exhibit distinct semantic tendencies: `really' conveys high Polarity and Surprisal, whereas `exactly' signals strong Polarity but low Surprisal. Both `mm' and `yeah' show weaker Polarity (with `yeah' slightly stronger), while `mm' generally expresses higher Surprisal than `yeah'. Despite these lexical tendencies, there is substantial variation driven by prosodic realization, leading to considerable overlap between lexical clusters. Although higher Surprisal and Polarity are broadly associated with longer durations and wider pitch ranges, Table~\ref{tab:affective_results} shows that simple prosodic features alone are insufficient predictors, suggesting that the relevant prosodic cues are more nuanced. A more detailed analysis of the central tendencies and dispersion of lexical forms with respect to the affective dimensions is provided in Appendix~\ref{sec:distribution}.

\section{Discussion}
\label{sec:discussion}

The findings in this work reinforce the view of backchannels as pragmatically conditioned signals whose lexical and prosodic forms are important for their appropriateness and nuanced meaning. The strong effect of context length suggests that backchannel choice depends on discourse-level structure and expectations rather than solely on local acoustic or lexical cues, and that such dependencies can be effectively captured by fine-tuned autoregressive language models.

The joint contrastive framework provides a useful abstraction for modeling backchannels in continuous space. Improvements over raw acoustic representations in perceptual similarity tasks indicate that the learned embedding space emphasizes pragmatically salient variation while down-weighting speaker- and token-specific idiosyncrasies. This is particularly evident in the cross-lexical similarity results, where different lexical forms are grouped according to perceived functional similarity rather than surface form alone.

As discussed earlier, while current generative SLMs should, in principle, be able to handle backchannels, they often struggle to generate and understand them correctly~\cite{lin2025full}. SLMs are trained to predict or synthesize the next segment of speech end-to-end, which entangles backchannel behavior with many other factors (lexical content, speaker identity, channel conditions) and makes fine-grained control of short feedback signals challenging. In contrast, we learn pragmatically meaningful representations in a joint embedding space that aligns dialogue contexts with backchannel realizations via a contrastive objective. This representation-centric formulation supports efficient retrieval/ranking, interpretable analysis (e.g., affective axes), and direct validation against human perceptual similarity — capabilities that are difficult to obtain from generation-focused objectives alone. Our approach could also be complementary to generative SLMs and could serve as a lightweight backchannel selection/control module within a more modular spoken dialogue system.

\section{Conclusion}
\label{sec:conclusion}

In this paper, we address the undermodeled link between the lexical–prosodic form of backchannels and their pragmatic appropriateness and interpretation in context. We introduced a two-stage approach that (i) fine-tunes an autoregressive LLM on spoken dialogue transcripts to obtain richer contextual representations and (ii) learns a shared embedding space that aligns dialogue contexts with acoustic backchannel realizations via contrastive learning. In retrieval-style evaluations, longer conversational context consistently improved performance, indicating that backchannel choice depends on pragmatics that extend beyond the immediately preceding turn.

Crucially, the learned projections produced backchannel embeddings that better match human perceptual structure than raw WavLM features: agreement increased in both prosodic and cross-lexical triadic similarity judgments. Linear probes further showed that the embedding space supports interpretable affective dimensions (Energy, Polarity, Surprisal), suggesting a path toward controllable feedback generation and pragmatic inference of user feedback in conversational systems.

\section*{Limitations}
\label{sec:limitations}

Modeling backchannels remains challenging due to their personality-dependent~\cite{warriner2013norms} and idiosyncratic nature~\cite{Blomsma_Vaitonyté_Skantze_Swerts_2024}, as well as their dependence on specific dyadic dynamics~\cite{cavalcanti2025dyadosyncrasy}. Furthermore, usage varies by language~\cite{heinz2003backchannel, bevnuvs2016prosody, liesenfeld2022bottom}, dialect~\cite{wong2007study, kraaz2022backchannels}, and speaker proficiency~\cite{cutrone2005case, shelley2013back, cutrone2014cross, lee2020backchannels, heinz2003backchannel}. In this work, we disregard these factors, experimenting with data collected from native speakers of United States English, irrespective of dialect. In the Fisher dataset, most participants did not know each other, making their dynamics more general and less clouded by previous acquaintance; however, we believe that future work needs to address individual, dyadic, and demographic differences to create more universally adaptive conversational agents.

While we have previously investigated the effect of different speech encoders \cite{liviaqa}, in this paper we consistently used WavLM over alternatives such as wav2vec 2.0 and HuBERT, due to its superior performance on the SUPERB benchmark~\cite{yang2021superb} and its robust capacity for modeling paralinguistic features~\cite{9814838}. We also kept the audio context length short (1 second), given that the simple temporal mean-pooling we used cannot capture  complex temporal dynamics or longer-term acoustic dependencies anyway. Furthermore, we kept the WavLM weights frozen throughout the training process to focus specifically on the alignment mechanism. While fine-tuning the audio backbone or employing different self-supervised speech models (or layers) might yield higher absolute performance metrics, we assume that the relative trends and the efficacy of the contrastive framework observed in this study would likely remain consistent.

Due to resource constraints, we fine-tuned smaller LLMs (4B and 7B), requiring 1-2 days on 8 NVIDIA RTX 3090 GPUs per model. We avoided LLM hyperparameter tuning, as the performance differences between model sizes were negligible. Focusing on context representations, we did not heavily optimize backchannel encoding; however, we found that a linear projection of audio-only encodings performed best.

\section*{Ethical considerations}

As far as we are aware, our methods do not have any harmful effects, biases, or risks, apart from what can generally be expected from machine learning-based models (such as introducing demographic or linguistic biases). We did not develop new foundation models, nor have we created large-scale datasets that can be mass-deployed to cause intentional harm.

For data collection, participants were informed that we do not intend to store personal or sensitive data and that the information they provide is anonymized. No ethics approval was needed.

All coding was done in Python using \href{https://pytorch.org}{Pytorch}, and pre-trained models were downloaded from ~\href{https://huggingface.co}{Hugging Face}. AI assistants (\href{https://chatgpt.com}{ChatGPT} and \href{https://gemini.google.com}{Gemini}) were used to correct grammar and reformulate sentences in the paper. Additionally, they were used for coding in some instances.

\section*{Acknowledgements}

This work was partially supported by the Wallenberg AI, Autonomous Systems and Software Program (WASP) funded by the Knut and Alice Wallenberg Foundation, as well as the Swedish Research Council (VR) project 2020-03812.

\section*{Code and other resources}

The code, the results of the perception study, and links to the model weights are available on GitHub~\footnote{\url{https://github.com/qianlivia/Aligning-Backchannel-and-Dialogue-Context-Representations}}.

\bibliography{main}

\clearpage

\appendix

\section{Transcripts for LLM fine-tuning}
\label{sec:appendix_transcripts}

\subsection{Spoken dialogue transcripts}

For LLM fine-tuning, we used unpunctuated lowercase transcripts based on the ones provided with the Fisher dataset. A set of special characters were used to indicate speaker shifts and overlapping speech, however these were treated as ordinary characters from the perspective of the language model, so that no new tokens had to be introduced: 

\begin{itemize}
    \item \textbf{<A>} and \textbf{<B>}: the beginning of Speaker A or B's turn, respectively
    \item \textbf{Slash} (\textbf{$/$}): turn shift
    \item \textbf{Braces} (\textbf{\{\}}): the part of a given turn (Turn A) that overlaps with parts of the next turn (Turn B). There is always a corresponding part surrounded by square brackets in the next turn (Turn B), unless the current turn (Turn A) is the last one in the transcript.
    \item \textbf{Square brackets} (\textbf{[]}): the part of a given turn (Turn B) that overlaps with parts of the previous turn (Turn A). There is always a corresponding part surrounded by braces in the previous turn (Turn A), unless the current turn (Turn B) is the first one in the transcript.
\end{itemize}

Backchannels were located by searching for complete turns that consisted \textit{only} of the backchannels we are investigating (see Section~\ref{sec:datasets}) and \textit{only once}, i.e., we identified them only if they were surrounded by two turn shifts and a speaker token (with occasional braces and brackets).

Here are two example transcripts, with backchannels marked in bold:

\subsection*{Example transcript 1}

\newenvironment{ttquote}
  {
    \par
    \addvspace{0.8\baselineskip}
    \noindent
    \ttfamily
    \footnotesize
    \setlength{\parindent}{0pt}
  }
  {
    \par
    \addvspace{0.8\baselineskip}
  }

\begin{ttquote}
    <B> hi / <A> hello / <B> hi / <A> oh okay um my name is brandon um you / <B> right um my name's rajul mm / <A> okay okay so uh what do you think / <B> um i didn't quite catch the topic i mean i got the gist of it but um can you hear the topic of the day / <A> okay basically um how has corporate scandals affected you or do you think um what is what is the affect of corporate scandals on america um do you believe it's responsible for the m the mild recession that we we been having lately mm / <B> \textbf{right} / <A> \textbf{mhm} / <B> um well uh personally for me to the the solution and uh you know the aftermath of the uh uh the uh tech boom and the uh bubble thereafter because i think um probably um because uh i think because of my ignorance at that point in time about / <A> mhm mhm / <B> um about probably what i would call uh the truth as i see it wh what actually goes on in in the um in in the stock market as well as uh uh the biggest uh financial institutions uh in america um but initially e i i i really could not believe uh um things that were coming out uh in in the newspapers you know arthur anderson and uh citibank merrill lynch you know just n name a all those groups were like uh um being charged for um um for misguiding and misleading uh the investors / <A> mhm mhm / <B> and and thereby um in my words uh duping people uh of their money uh uh pretty blatantly \{ knowing \} / <A> [ \textbf{mhm} ] / 
\end{ttquote}

\subsection*{Example transcript 2}

\begin{ttquote}
    <A> and you know we we uh all / <B> \textbf{yeah} / <A> of us hurt uh because of that and i believe it's contributed to the recession it's i mean the recess recessions happen because they happen i mean they they come in waves but / <B> \textbf{right} / <A> i don't i don't think that me i don't think that america takes white collar white collar crimes seriously and \{ it \} / <B> [ \textbf{right} ] / <A> doesn't it doesn't try to it doesn't try to stop it at all i don't i think i don't know \{ h \} / <B> [ \textbf{right} ] / <A> uh if you have a theory behind that because also i think the main people who are running the / <B> \textbf{mhm} / <A> the country the main people who are actually contributing to politicians who run the countr the country are all these companies \{ who are \} / <B> [ \textbf{right} ] / <A> so rich and have all this money and so if they you know if they embezzle a few million dollars it doesn't hurt anyone cause we're still in power \{ th \} / <B> [ \textbf{right} ] / <A> these are the um the people in power who are talking you know we're st we're still in power and like they're still paying us off so it doesn't really matter but everyone \{ but \} / <B> [ right right ] / <A> everyone else hurts in the long run so i don't know /
\end{ttquote}

In \textit{Example transcript 1}, the penultimate turn overlaps with the last turn, which means that \textbf{\{ knowing \}} is uttered at the same time as \textbf{[ mhm ]} in the original audio. Other vocalized dialogue phenomena are included, e.g., repetitions and repairs, without special markers.

\subsection{Defining the context}

When restricting the number of past turns to, e.g., 2, the contexts corresponding to some of the above examples appear as follows (where everything before the backchannel in bold is treated as context):

\begin{ttquote}
    <B> um i didn't quite catch the topic i mean i got the gist of it but um can you hear the topic of the day / <A> okay basically um how has corporate scandals affected you or do you think um what is what is the affect of corporate scandals on america um do you believe it's responsible for the m the mild recession that we we been having lately mm / <B> \textbf{right}
\end{ttquote}

\begin{ttquote}
    <A> okay basically um how has corporate scandals affected you or do you think um what is what is the affect of corporate scandals on america um do you believe it's responsible for the m the mild recession that we we been having lately mm / <B> right / <A> \textbf{mhm}
\end{ttquote}

\begin{ttquote}
    <A> mhm mhm / <B> and and thereby um in my words uh duping people uh of their money uh uh pretty blatantly \{ knowing \} / <A> [ \textbf{mhm} 
\end{ttquote}

Thus, when computing the perplexity of the backchannel token, or the LLM context embedding, all text up to the backchannel (marked in bold) is used, including the current turn's speaker token and potential opening brackets.

\section{Best hyperparameters}
\label{sec:hyperparam}

 The hyperparameters of the best models for each configuration in Section~\ref{sec:joint_learning} are shown in Table~\ref{table:hyperparam1} and~\ref{table:hyperparam2}.

\begin{table}[t]
\centering
\begin{tabular}{lcc}
\toprule
& \multicolumn{2}{c}{\textbf{Context modality}} \\
\cmidrule(lr){2-3}
\textbf{\makecell[l]{Text \\ embeddings}} & Audio + text & Text \\
\midrule
Gemma 3 4B   & 4, 128, 2048 & 2, 128, 4096 \\
LLaMA 3.1 8B &  3, 64, 4096 & 3, 128, 4096 \\
Qwen2.5 7B   &  4, 64, 2048 & 1, 128, 2048 \\
Mistral 7B   &  3, 64, 2048 &  1, 64, 8192 \\
\bottomrule
\end{tabular}
\caption{Hyperparameters for the models (5 turns) in Table~\ref{tab:joint}, listed in the order \textit{number of layers}, \textit{embedding size}, \textit{batch size}. For audio-only, the optimal configuration was 2, 64, and 2048.}
\label{table:hyperparam1}
\end{table}

\begin{table}[t]
\centering
\begin{tabular}{lcc}
\toprule
 Model       & 1    & 3       \\
\midrule
Gemma 3 4B   &  2, 64, 2048 & 3, 64, 1024 \\
LLaMA 3.1 8B & 4, 256, 2048 & 2, 64, 4096 \\
Qwen2.5 7B   &  3, 64, 2048 & 3, 64, 2048 \\
Mistral 7B   &  3, 64, 2048 & 4, 64, 4096 \\
\bottomrule
\end{tabular}
\caption{Hyperparameters for the models (text + audio, fine-tuned LLM context encoder) in Table~\ref{tab:joint-ctxlen} in the order \textit{number of layers}, \textit{embedding size}, and \textit{batch size}. The number of past turns is 1 and 3.}
\label{table:hyperparam2}
\end{table}

\section{Perception study}
\label{sec:appendix_questions}

In the perception study, participants were given three tasks, with the last one consisting of three separate questions regarding Energy, Polarity and Surprisal. In total, five questions were asked for each stimulus set. The participants were shown a range of examples at the beginning of the study.

The perception study was conducted on 100 native speakers of North American (U.S.) English who reported English as their primary — most frequently used — language and the United States as their primary place of residence during their first 18 years. The participants had no hearing difficulties. Each participant received 21 stimulus sets and two additional sets for attention checks. Each stimulus set was seen by at least three subjects.

Participants were recruited via \href{https://www.prolific.com}{Prolific} and the experiment was hosted on \href{https://www.cognition.run}{cognition.run}. The median completion time was approximately 56 minutes, and the participants were compensated with Prolific's default reward per hour.

\section{Outline of perception study questions}
\label{sec:appendix_questions2}

\paragraph{General instruction:} Listen to the context and the feedback responses. The feedback responses are repeated at the beginning of each question where you have to rate them individually.

[context audio]

[context transcript]

[Feedback 1, Feedback 2, Feedback 3 (only audio)]

\paragraph{Question 1:} Please rate the feedback responses based on how well they match the context (min: 1, max: 5).

\paragraph{Question 2:} Choose the two responses that are the most similar to each other.

\paragraph{Question 3:} Rate the energy level: how energetic is the response (min: 1, max: 5)?

\paragraph{Question 4:} Rate polarity: how positive is the response (min: 1, max: 5)?

\paragraph{Question 5:} Rate surprisal: how surprised does the feedback speaker sound (min: 1, max: 5)?

\section{Informed consent and general information (verbatim)}
\label{sec:informed}

For each question, you will hear:
\begin{itemize}
    \item one \textbf{context} clip (Speaker 1)
    \item three short \textbf{feedback} clips (Speaker 2), labeled as 1, 2 and 3
\end{itemize}

\subsection*{Your task}

\begin{enumerate}
    \item \textbf{Rate compatibility}
    \begin{itemize}
        \item For each feedback, rate how well it fits the context as a possible response
        \item Scale: $1 =$ not at all, $5 =$ extremely
        \item Consider factors like naturalness, appropriateness and expectedness
    \end{itemize}
    
    \item \textbf{Choose similar feedback responses}
    \begin{itemize}
        \item Pick the two feedback responses most similar in role, attitude, intention and emotion (not just meaning!), given the current context
        \item Options: 1--2, 2--3 or 1--3
        \item Consider factors like how replaceable they are with each other! Take into account both sound and meaning.
    \end{itemize}
    
    \item \textbf{Rate energy}
    \begin{itemize}
        \item How energetic does the feedback speaker sound?
        \item Scale: $1 =$ not at all, $5 =$ extremely
    \end{itemize}
    
    \item \textbf{Rate polarity}
    \begin{itemize}
        \item How positive is what the feedback speaker said?
        \item Scale: $1 =$ not at all, $5 =$ extremely
    \end{itemize}
    
    \item \textbf{Rate surprisal}
    \begin{itemize}
        \item How surprised does the feedback speaker sound?
        \item Scale: $1 =$ not at all, $5 =$ extremely
    \end{itemize}
\end{enumerate}

\subsection*{Instructions}

\begin{itemize}
    \item You will see some examples on the first six pages. Please, read them carefully.
    \item \textbf{You have to wait a few seconds for the survey to load.}
    \item \textbf{Evaluate feedback together with context}, not in isolation. For better judgment, play each feedback \textbf{immediately after the context}.
    \item Consider tone, emotion, and purpose --- but \textbf{ignore} differences in recording quality and loudness.
\end{itemize}

\hrulefill

\subsection*{Listening requirements}
\begin{itemize}
    \item Use \textbf{headphones} in a calm environment.
    \item Replay samples as needed, but you \textbf{cannot return} to earlier ones.
    \item The volume of the context can be adjusted.
    \item If an audio file doesn't load or doesn't play, wait a few seconds. Once loaded, it should be possible to play it without delay. If it still does not work, please take a note of where the issue arose and contact us.
\end{itemize}

\subsection*{Attention checks}
\begin{itemize}
    \item Very short or long completion times may lead to rejection.
    \item Failing attention checks results in rejection. Listening activity \textit{is recorded} for validation.
    \item The survey will not end if you fail the attention checks, and your answers, including the checks, will be evaluated after your submission.
\end{itemize}

\subsection*{Other notes}
\begin{itemize}
    \item Answers will help study how people perceive \textit{backchannels} in conversation.
    \item Audio clips are from a dataset and do not represent our views.
    \item Responses are anonymized; no personal/sensitive data is collected.
    \item You may stop anytime or contact us via Prolific if needed.
\end{itemize}

\noindent \textbf{By continuing, you confirm voluntary participation and acceptance of these conditions.}

\vspace{1em}
\noindent Thank you!

\section{Distribution of affective ratings}
\label{sec:distribution}

The mean and standard deviation over the median scores for each backchannel, grouped by lexical category, are shown in Table~\ref{tab:triad_pm}. These provide insight into the typical human perception of each category and how it varies within each group. The data suggests a hierarchy from high-arousal, expressive words (top of the table) to low-arousal, passive continuers (bottom of the table).

\paragraph{Highest Energy:} The token ``wow'' ($3.74 \pm 0.70$) has the highest average Energy, indicating that it is perceived as the most intense or aroused response in general. It is closely followed by strong, occasionally enthusiastic agreements like ``absolutely'' ($3.67$) and ``exactly'' ($3.67$).

\paragraph{Lowest Energy:} The tokens ``mhm'' ($2.70 \pm 0.68$), ``yeah'' ($2.83$), and ``right'' ($2.87$) are at the bottom. This aligns with their typical linguistic function as ``passive'' continuers, i.e., simple signals that the listener is still present, without adding significant emotion.

\paragraph{Most positive:} ``Absolutely'' ($4.03 \pm 0.67$) and ``definitely'' ($3.98 \pm 0.88$) are the most positive tokens. This suggests that multi-syllabic, explicit agreement words carry more positive weight than short sounds.

\paragraph{Least positive (most neutral):} The lowest Polarity scores belong to ``mm'' ($2.71$) and ``mhm'' ($2.92$). These scores are likely lower not because they are negative, but because they are highly neutral (our selection does not, for the most part, include backchannels that are usually perceived as expressing negative sentiment).

\paragraph{Most surprised:} Words like ``wow'' ($3.92$), ``really'' ($3.84$), ``ah'' ($3.82$), and ``oh'' ($3.69$) all have very high Surprisal scores. These are typically used when the listener is reacting to new, shocking, or interesting information.

\paragraph{Least surprised:} Words like ``mhm'' ($2.04$), ``yep'' ($2.21$), ``right'' ($2.22$), and ``sure'' ($2.30$) have low Surprisal scores. These are used to confirm known information or simply to agree.

\paragraph{High variance/variability:} Certain tokens have consistently high standard deviations, suggesting that they can be pronounced in various ways and that their meaning depends heavily on how they are said (tone/prosody) rather than just on the word itself. Examples include ``oh'', ``mm'', and ``cool''.

\begin{table*}[t]
\caption{Mean ± standard deviation per backchannel lexical token. Calculated on the median rating of each backchannel.}
\label{tab:triad_pm}
\begin{tabular*}{\textwidth}{@{\extracolsep{\fill}} lccc}
\toprule
token & energy & polarity & surprisal \\
\midrule
wow & 3.74 $\pm$ 0.70 & 3.64 $\pm$ 0.61 & 3.92 $\pm$ 0.71 \\
absolutely & 3.67 $\pm$ 0.73 & 4.03 $\pm$ 0.67 & 2.45 $\pm$ 0.56 \\
exactly & 3.67 $\pm$ 0.61 & 3.96 $\pm$ 0.57 & 2.41 $\pm$ 0.66 \\
ah & 3.57 $\pm$ 0.74 & 3.27 $\pm$ 0.67 & 3.82 $\pm$ 0.78 \\
really & 3.56 $\pm$ 0.74 & 3.22 $\pm$ 0.72 & 3.84 $\pm$ 0.74 \\
definitely & 3.51 $\pm$ 0.70 & 3.98 $\pm$ 0.88 & 2.34 $\pm$ 0.44 \\
oh & 3.45 $\pm$ 0.86 & 3.17 $\pm$ 0.87 & 3.69 $\pm$ 0.89 \\
good & 3.34 $\pm$ 0.71 & 3.55 $\pm$ 0.59 & 2.84 $\pm$ 0.63 \\
yes & 3.11 $\pm$ 0.73 & 3.39 $\pm$ 0.65 & 2.21 $\pm$ 0.62 \\
cool & 3.07 $\pm$ 0.83 & 3.49 $\pm$ 0.69 & 2.65 $\pm$ 0.81 \\
okay & 3.02 $\pm$ 0.64 & 3.21 $\pm$ 0.58 & 2.54 $\pm$ 0.69 \\
uh-huh & 3.00 $\pm$ 0.65 & 3.07 $\pm$ 0.69 & 2.38 $\pm$ 0.67 \\
yep & 2.95 $\pm$ 0.62 & 3.30 $\pm$ 0.57 & 2.21 $\pm$ 0.61 \\
sure & 2.89 $\pm$ 0.71 & 3.34 $\pm$ 0.64 & 2.30 $\pm$ 0.71 \\
mm & 2.88 $\pm$ 0.82 & 2.71 $\pm$ 0.70 & 2.82 $\pm$ 0.95 \\
right & 2.87 $\pm$ 0.66 & 3.30 $\pm$ 0.66 & 2.22 $\pm$ 0.52 \\
yeah & 2.83 $\pm$ 0.62 & 3.14 $\pm$ 0.55 & 2.20 $\pm$ 0.60 \\
mhm & 2.70 $\pm$ 0.68 & 2.92 $\pm$ 0.61 & 2.04 $\pm$ 0.58 \\
\bottomrule
\end{tabular*}
\end{table*}

\end{document}